# Bubble identification from images with machine learning methods

H. Hessenkemper[1], S. Starke[2], Y. Atassi[1], T. Ziegenhein[1,3], D. Lucas[1]

[1]Helmholtz-Zentrum Dresden-Rossendorf, Institute of Fluid Dynamics, Bautzner Landstraße 400, 01328 Dresden, Germany

[2]Helmholtz-Zentrum Dresden-Rossendorf, Department of Information Services and Computing, Bautzner Landstraße 400, 01328 Dresden, Germany

[3]TIVConsultancy, 7th Pl 1312, Tempe, AZ 85281, USA

\* Corresponding author. Tel.: +49 3512604719; fax: +49 3512603440.
E-mail address: h.hessenkemper@hzdr.de (Hendrik Hessenkemper)

**Abstract**

An automated and reliable processing of bubbly flow images is highly needed to analyse large data sets of comprehensive experimental series. A particular difficulty arises due to overlapping bubble projections in recorded images, which highly complicates the identification of individual bubbles. Recent approaches focus on the use of deep learning algorithms for this task and have already proven the high potential of such techniques. The main difficulties are the capability to handle different image conditions, higher gas volume fractions and a proper reconstruction of the hidden segment of a partly occluded bubble. In the present work, we try to tackle these points by testing three different methods based on Convolutional Neural Networks (CNN's) for the two former and two individual approaches that can be used subsequently to address the latter. Our focus is hereby on spherical, ellipsoidal and wobbling bubbles, which are typically encountered in air-water bubbly flows. To validate our methodology, we created test data sets with synthetic images that further demonstrate the capabilities as well as limitations of our combined approach. The generated data, code and trained models are made accessible to facilitate the use as well as further developments in the research field of bubble recognition in experimental images.

**Keywords.**   Bubbly flows, Deep Learning, Computer Vision, CNN, Instance segmentation

# 1. Introduction

Deep neural networks have proven their superiority over traditional computer vision methods in various fields. Especially Convolutional Neural Networks (CNNs) have been shown to be very successful for image segmentation tasks (He et al., 2020; Ronneberger et al., 2015; Schmidt et al., 2018). They can achieve a high segmentation resolution, making the direct use of such methods interesting for investigating bubbly flows.

Using cameras to investigate bubbly flows from the outside is a common, affordable technique. However, the images' evaluation can be very complicated when bubbles start to overlap, strong turbulences are formed, and dense bubble swarms occur. Yucheng Fu provides an in-depth discussion of this problem in his dissertation (Fu, 2018).

The task of recognizing bubbles with CNNs from images can be usually split up into identifying the bubble(s) in the picture, segmentation of overlapping bubbles and reconstruction of bubbles that are partly occluded from bubbles before. A common way to solve the task of identifying bubbles is to find so-called anchor points inside the bubble. Ideally, just one point per object exists, which is usually the center point. With respect to machine learning approaches, Haas et al., (2020) use a popular Region-based CNN called Faster-RCNN that proposes anchor points with corresponding bounding boxes around identified bubbles. Poletaev et al., (2020) use a CNN-based sliding window approach to approximate anchor points. Another approach to solve this task together with the task of segmenting overlapping bubbles is to directly predict a segmentation mask for an image with down- and upsampling CNN's that classify and assign each pixel to individual objects. Such pixel-to-pixel approaches have become very popular for detecting and segmenting cells and nuclei in biomedical microscopic images. For bubbly flows, Li et al., (2021) used a UNet to distinguish between foreground and background pixels as well as to generate centroid approximations. Kim and Park, (2021) used a slightly customized Mask-RCNN version that directly provides a segmentation mask as a result.

For most of the above-mentioned studies on identifying bubbles, an ellipsoidal fit segments and reconstructs the objects solving the tasks of segmenting and reconstructing overlapping bubbles. Haas et al., (2020) used a subsequent CNN-based shape regression model that tries to fit an ellipse around the detected bubble. Cerqueira and Paladino, (2021) also used a CNN-based shape estimator to predict ellipses for given anchor points and bounding boxes. From our experience, an ellipsoidal fit is, however, only valid in simplified bubbly flows. With flow fields disturbed by turbulence and swarm effects, the surface tension may not be strong enough in relation to deforming forces so that bubbles can take almost arbitrary shapes (Masuk et al., 2021).

In this work, we also adopt the strategy to use CNN's that provide pixel-to-pixel predictions in order to identify and segment bubbles. In particular, we test three different approaches for this. For the subsequent reconstruction task, we compare two individual approaches, one also based on fitting an ellipse and one based on a simple Neural Network. The latter aims to capture the hidden part of a partially occluded bubble even with more irregular shapes, allowing a more universal use of this approach. Finally, we test the accuracy of our combined approach on synthetic images with known gas volume fraction and bubble sizes to evaluate strength and weaknesses.

# 2. Methods & Materials

## 2.1. Bubble segmentation

### *Double UNet*

In the following, the methods tested in this work for bubble identification and segmentation are described. The first method is based on so-called UNet's. A UNet describes a specific CNN architecture that allows a pixel-to-pixel prediction, which is commonly used in segmentation tasks for cell structures and medical images (Ronneberger et al., 2015). The name originates from the U-like shape of the CNN, consisting of an encoder and a decoder structure that have cross-connections between them. The encoder structure reduces the width and height of an array but increases the depth (channels), to extract features from an image, while the decoder structure does the opposite in order to obtain local information in the image. The general UNet architecture together with the parameters used in this work are shown in Figure 1.

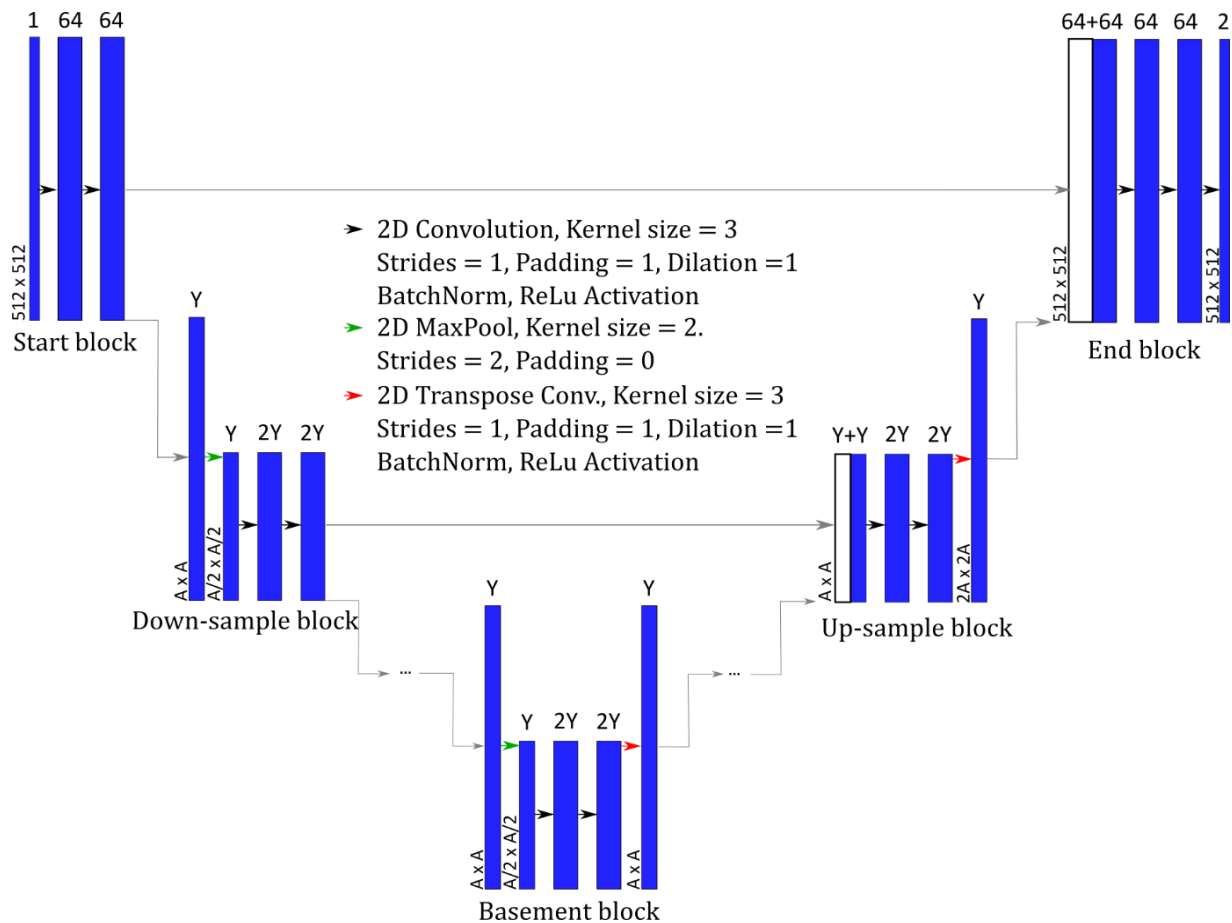

Figure 1: UNet architecture and used parameters. Y refers to the number of filters and A to the input dimension for the respective layer, where Y increases and A decreases up to the basement block. Correspondingly, the Basement block of UNetL3 has Y=256 and A=128, while the UNetL5 has Y=1024 and A=32.

In order to segment bubbles in images, we use two individual UNet's with slight modifications in comparison to the original one by Ronneberger et al., (2015). The first UNet (UNetL3) consists of three down- and upsampling levels and is trained to distinguish between foreground and background, in other words to classify all pixels whether they belong to the gas phase or to the liquid phase. The second UNet (UNetL5) is trained to classify all pixels that belong to intersections of overlapping bubbles. Since this task is more difficult, a deeper net with five down- and upsampling levels is used. For this UNetL5, we use manually annotated edges (Figure 2) to calculate the loss function of the output of the network. In the context of machine learning, the loss function determines the error of the algorithm with respect to the target result, which is then minimized in the training step. For the UNetL5, the target is the correct classification of all pixels marked in red in Figure 2 as intersecting pixels. We use a log softmax cross entropy loss function to achieve this classification. Since the edge pixels are under-represented compared to the background pixels, we weight the background pixels' loss, so that the training focuses on the intersection pixels the network is intended to find. Finally, we calculate the weighted average of the loss of all pixels. We tested networks with fewer layers, with however limited success, the sigmoid cross entropy loss function with one output, which performs similar to the log softmax cross entropy and different weight factors for the background loss. For the latter, a value of 0.05 provided the best results.

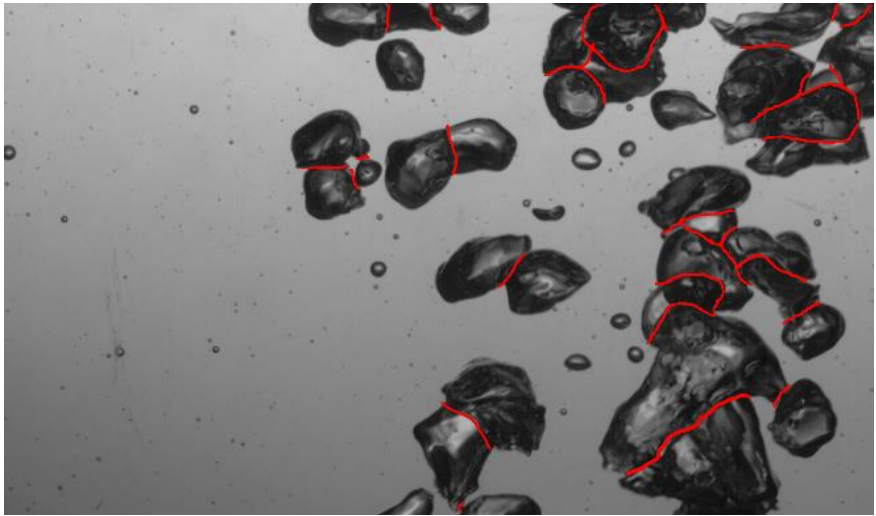

Figure 2: Marked internal edges the UNetL5 is trained on, which are used to segment overlapping bubbles.

For the UNetL3 the loss function is similar to the one we use for the edges; however, instead of just applying a weighting-factor, a weight map is used to focus the learning on specific structures. Using a weight-map is necessary so that the network training is focused on separating bubbles that have only a small gap of background pixels in between them. The weight map is calculated as follows:

$$w = \begin{cases} 10, & d_1 \text{ and } d_2 < 10 \\ 1, & \text{inside bubble} \\ 0.05, & \text{else} \end{cases}$$

where $d_1$ and $d_2$ are the distances between one interface and another interface. The weight map to train the segmentation between bubbles and background for the example given in Figure 2 is shown in Figure 3.

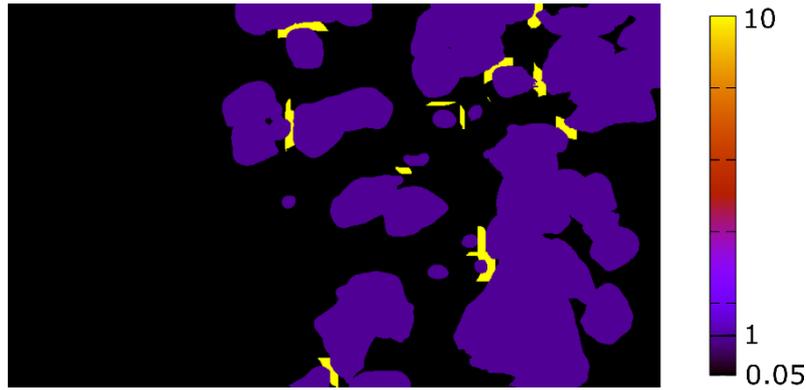

Figure 3: Weight map for the loss function to train the segmentation between bubbles and background (UNetL3) for the example given in Figure 2.

A further advantage of having two individual UNet's is that the mask generated by the smaller and hence faster UNetL3 can also be used in PIV/PSV investigations to exclude bubbles in the liquid velocity interrogation step (Cerqueira et al., 2018; Hessenkemper and Ziegenhein, 2018).

### *StarDist*

The second method tested is called StarDist. It was also initially developed for segmenting cell nuclei in biomedical images by proposing star-convex polygons as object candidates (Schmidt et al., 2018). Since the shape of many bubbles can be well approximated with such star-convex polygons, we adopt this method without further modification. As for the first method, StarDist is also based on a UNet architecture for pixel-to-pixel predictions, but with a more sophisticated strategy. In particular, StarDist generates two arrays, both having the dimension of the input image. The first output is an object probability $d_{i,j}$ for each pixel defined as the normalized Euclidean distance to the nearest background pixel. Hence, this output is similar to distinguishing foreground-background like it is done with the UNetL3, but with continuous values that reach higher values in the center of the mask, which then have a higher probability to serve as object centers. The second output represents the Euclidean distance $r_{i,j}^k$ to the background for every pixel belonging to an object along a fixed set of $k$ radial directions. In other words, for every object pixel a star-convex polygon with $k$ points is proposed. Finally, by applying non-maximum suppression (NMS), only pixels with a high object probability are considered to avoid detecting a single instance multiple times. Figure 4 shows examples of star-convex polygons representing the outline of a bubble with a fixed number of $k = 64$ radial directions. We tested different hyperparameters such as more UNet layers, different subsampling resolutions to increase computation efficiency as well as different number of radial directions. By comparing the obtained results (see section 3.1 for the evaluation strategy) we found that a three layer UNet with $k = 64$ radial directions provided the best results for our task, while keeping all other hyperparameters as in the original version of Schmidt et al.,( 2018).

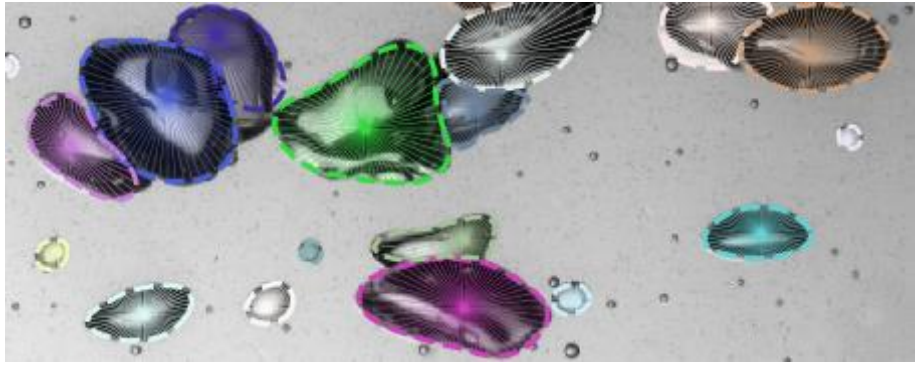

Figure 4: Star-convex polygons with radial distances applied on a bubbly flow image.

*Mask R-CNN*

The third method is called Mask R-CNN. It has been used for numerous image segmentation tasks including also the segmentation of overlapping bubbles as done recently by Kim and Park, (2021). Mask R-CNN consists of an object detector based on the Faster R-CNN method (Ren et al., 2017), which predicts bounding boxes around found objects. In parallel, Mask R-CNN creates individual segmentation masks for each region of interest (ROI) with a Fully Convolutional Network, which again gives pixel-to-pixel results. As Kim and Park, (2021) already tested this method for the task of segmenting overlapping bubbles and obtained quite good results with some modifications in comparison to the original Mask R-CNN version by He et al., (2020), we adopt their version of Mask R-CNN with only minor parameter changes. Here we only increased the minimum detection confidence to 0.9, which is the probability threshold for detected instances and increased the number of training epochs to 40. For the reason stated above, the Mask R-CNN results are only used for comparison and are not discussed in detail.

*Training*

The training data set consists of roughly 800 manually annotated training images with dimensions of 512 x 512 (width x height). Since our goal is to obtain a rather universal model that can be applied to different kinds of bubbly flow images, we have conducted experiments with various flow and imaging conditions, i.e. different cameras, lenses, camera distances and illuminations. Most of the experiments consider air-water flows in buoyancy driven bubbly flows with gas flow rates in the range of 0.5 - 2 l/min and a corresponding gas fraction in the range of 1 – 5 %, together with some experiments in water-glycerol mixtures with a logarithmic Morton number of -6.6. The dataset consists of about 24,400 individual annotated bubbles in the size range of 1 – 10 mm with dominantly spherical, ellipsoidal and wobbling bubble shapes. More information on the facilities in which the experiments were conducted can be found in (Liu et al., 2019; Ziegenhein and Lucas, 2019).

To prevent overfitting, we apply image normalization and further random image augmentation steps to the training images, including adding Gaussian noise, intensity changes and horizontal flipping. We excluded 15 % of the training data from the training as validation data, for which the performance details will be discussed in section 3.1. For the Mask R-CNN approach, we used the ImageNet pre-trained weights, while no pre-trained weights were used for the other models. The training was performed on a NVIDIA

Tesla V100. Additionally, the following Table 1 lists the computational aspects regarding training and inference of the tested models.

Table 1: Computational aspects regarding training and inference of the different models.

| Name | No. of training epochs [-] | Time per epoch [s] | Inference time CPU [s] |
|---|---|---|---|
| UNetL3 | 100 | 212 | 1.6 |
| UNetL5 | 100 | 370 | 4.12 |
| StarDist | 400 | 29.9 | 0.51 |
| Mask R-CNN | 40 | 696 | 6.5 |

## 2.2. Hidden part reconstruction

As will be shown and discussed in the result section, all of the above described methods can be used to segment overlapping bubbles, but are not able to reconstruct the hidden part of a partly occluded bubble. In other words, with such a segmentation it is only possible to separate the visible parts, which would result in an underprediction of the partly occluded bubble. Based on a given segmentation mask we test two different methods to overcome this shortcoming.

### *Ellipse fitting*

The first method is comparably simple and follows the often used approach to fit an ellipse around the contour of a detected bubble. This, however, does not help to reconstruct the hidden part of a bubble when using only the detected segments. For this reason, only the contour points of a segment that do not exhibit a neighboring contour point of another segment are used for the ellipse fitting step. The two left images of Figure 5 show an example for this method. In cases of highly occluded bubbles, with only a few contour points that do not touch neighboring segments, this can lead to very small ellipses that are smaller than the actual segment. Then an ellipse is fitted around the complete segment to ensure that the fitted ellipse is at least as large as the detected segment.

### *Radial distance correction (RDC)*

The second method follows the algorithm idea of StarDist to represent the contour of an object through a fixed number of $K$ radial distances $r = (r_1, \ldots, r_K)$ from the object center to the boundary. In particular, the radial distances of a partly occluded bubble $r^H = (r_1^H, \ldots, r_K^H)$ are ending at the segmentation boundary and are hence shorter than for the actual object: $r_i^H \leq r_i, i = 1, \ldots, K$. It therefore requires a function to properly extend those directions $r_i^H$ that touch a neighboring bubble based on the information given by the image segment, in other words based on the not hidden radial distances. In particular, we want a function that is able to predict the correct distances of the occluded bubble part to the center of the detected segment, which will be called radial distance correction (RDC) in the following. This idea is illustrated in the two right images of Figure 5, where the red colored radial distances are the shortened ones that need to be corrected.

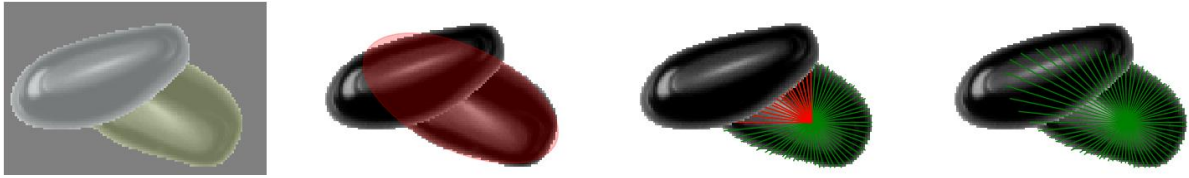

Figure 5: Principles of the tested methods to reconstruct hidden bubble parts. From left to right: Segmentation mask, fitted ellipse, radial distances and ground truth radial distances.

To solve this regression task, we again make use of a Neural Network, but this time we use a feedforward artificial neural network. The input and output layer represent the fixed number of radial distances $r$, in our case $K = 64$. Hence, both layers consist of 64 neurons. We use three hidden layers with the same number of neurons as the input and output layer, both with a ReLu activation function. Furthermore, we use an Adam optimizer with a learning rate of $10^{-4}$.

### *Training*

In order to train the RDC method, it is necessary to know the actual contour of a partly occluded bubble as ground truth. This is only possible with synthetic images, in other words images where bubbles are artificially placed on top of each other with respect to the view direction. Although this is possible with a generative adversarial network (GAN) (Fu and Liu, 2019), we use images of single bubbles that are cut out along their contour as done in our previous investigation (Hessenkemper et al., 2021). About 14,000 single bubble images in the size range of 2 – 7 mm were used to create the synthetic images, where either two or three bubbles were placed randomly on top of each other in an empty image with dimensions of 256 x 256. In total, about 150,000 training samples were created, where the radial distances $r$ are generated with the ideal segmentation mask (Figure 5, leftmost image). Ideal segmentation refers hereby to the correct assignment of each individual pixel to the correct bubble instance, which is given due to the synthetic nature of the data. The visible $r$ values of each bubble (Figure 5, third image from left) as well as the corresponding ground truth $r$ values (Figure 5, rightmost image) were extracted for the training. Care was taken that the bubbles overlap each other with at least 10 % but not more than 90 % of their area to avoid a complete occlusion. To be able to learn whether $r^H$ belongs to an occluded bubble that needs correction or whether $r^H$ belongs to a bubble in front that does not need correction, also the $r^H$ distances that do not need correction are included in the training. A further important point is that we do not use the initial pixel distances, but scale them with the known physical pixel size to obtain real distances. The advantage is that the model implicitly incorporates the size dependent shape of the bubble and is furthermore independent of the resolution of the image. Training was performed on 32 CPU cores with 2000 epochs and a batch size of 1400. Again, a small part of the data were split up before the training as validation data. Due to the larger data set size, we use only 6.67 % (10,000 samples) as validation data.

# 3. Results

## 3.1. Bubble segmentation

*Validation results*

In order to evaluate the performance of the segmentation task, we use the Average Precision (AP) value at Intersection over Union (IoU) thresholds from 0.5 to 0.9. The Average Precision is commonly used when evaluating classification tasks and combines the metrics Precision and Recall, where the former focuses on the proportion of predictions that are actually correct, while the latter focuses on the proportion of actual instances that were correctly determined (Zhang and Su, 2012). The IoU thresholds determine how much of a segment needs to be captured correctly, i.e. for high IoU thresholds the predicted pixel area of an segment needs to be close to the area of the actual segment (in size and position) to be counted as correctly predicted and vice versa for low IoU thresholds.

We first test the different methods on the validation data, which is a subset of the training data that has not been used in the training. All three methods show good results, which demonstrates the general capability to detect and segment overlapping bubbles (Figure 6). For all IoU thresholds the UNet method shows the best result, followed by the StarDist method. This is further illustrated with the example images in Appendix A. As can be seen, the main advantage of the UNet method is that it is especially good in catching the whole bubble area, while StarDist misses some of the outer bubble regions. In Figure 6 this is reflected in higher AP scores of the UNet method at higher IoU thresholds. The Mask R-CNN results are close to the StarDist results, with slightly lower AP scores at higher IoU thresholds.

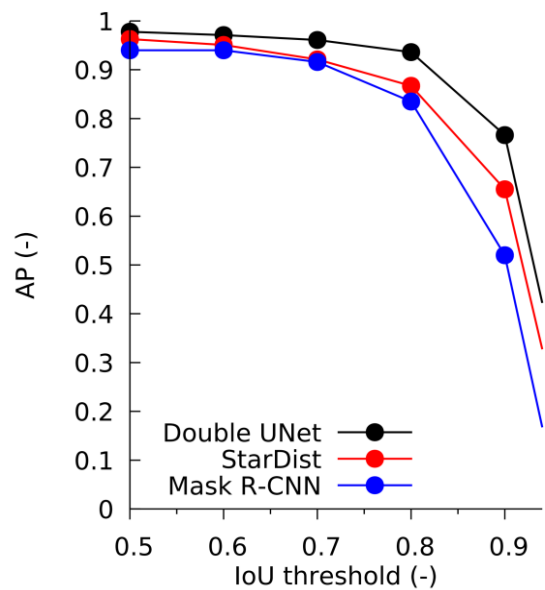

Figure 6: Results of the tested methods applied on the validation data set.

*Validation for different image conditions*

For more rigorous testing of the methods and to test the generalization ability a further performance evaluation with additional validation data is conducted in the following. Here we use images where the bubble projections show substantial differences to the training data as well as some more challenging cases, where more and also more deformed bubbles are present. Examples are shown in Appendix B. As can be seen, the UNet method fails to correctly predict the intersections here, with many incomplete and spurious intersections. This shows a conceptual drawback of the Double UNet method, as such intersections of overlapping bubbles are in some cases hardly visible or may look very similar to other internal edges due to distortions. However, the bubble masks provided by the UNetL3 still accurately catch the outline of the bubbles. On the other hand, the StarDist method is able to detect most of the bubbles but again misses the correct outline especially for larger bubbles. For both methods the described strengths and weaknesses are reflected in corresponding AP scores at given IoU thresholds, which are shown in Figure 7. To use the advantages of both methods, we combine the prediction results and dilate the StarDist results until it fits the bubble mask of the UNetL3. With this combination, we obtain better results than with the three other methods, which is reflected in the best AP scores in Figure 7 and can be further seen in Appendix B. Again, the Mask R-CNN achieves similar results as StarDist.

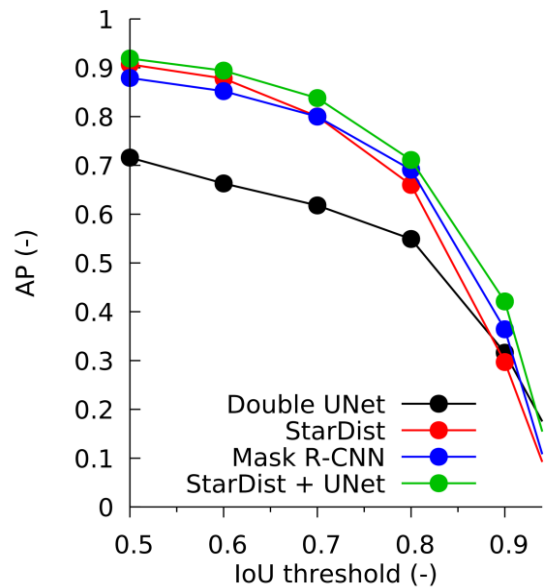

Figure 7: Results of the tested methods applied on the additional test data set.

### 3.2. Hidden part reconstruction and combined validation

To compare the results of fitting ellipses to the RDC method, we evaluate the captured area by each method with respect to the actual area of the ground truth bubble. Specifically, we calculate the root mean squared error (RMSE) of all bubbles in the validation data set. With respect to the average ground truth area, we get a relative error of 17 % with the ellipse fitting method and 11 % relative error with the RDC method. Although both methods provide similar results for the majority of the tested samples, the ellipse fitting method sometimes creates way too large predictions, which at the end cause the worse result in comparison to the RDC method. In order to further quantify the accuracy of our methods in practical situations, we generated test data sets with different gas volume fractions and applied the StarDist+UNet (SD+UNet) as well as the two hidden part reconstructions methods on them. We again use synthetic images for this to know the actual size of partly occluded bubbles, but this time we tried to design some more realistic images. Specifically, we use an image from a single bubble study without a bubble in it as background and then successively place single bubbles at random positions on it. The only constraint is that for every bubble at least 10 % of its area has to be visible, otherwise they can not be detected. Note that we use bubbles from another experimental series, which means the RDC training has not seen any of these bubbles. An example synthetic image is given in Figure 8 together with the segmentation mask predicted by SD+UNet and the RDC corrected bubble outline. To compare parameters relevant for practical uses of the methods, we calculate the volume of each bubble in order to determine a total gas fraction for every generated image. We set the dimensions of our artificial domain to have the height and width of our image and a depth of 3 cm. This depth was chosen because it allows the assumption that enough space is available for the overlapping bubbles, but no longer enough to allow a large number of fully occluded bubbles behind the visible bubbles. Furthermore, it reflects a typical depth dimension for experimental bubble columns or pipes (Ferreira et al., 2012; Hosokawa and Tomiyama, 2004; Lau et al., 2013; Pfleger et al., 1999). With this procedure, we generate test data for gas volume fractions of about 2.5 %, 5 %, 7.5 % and 10 %, where 50 images are generated for each case.

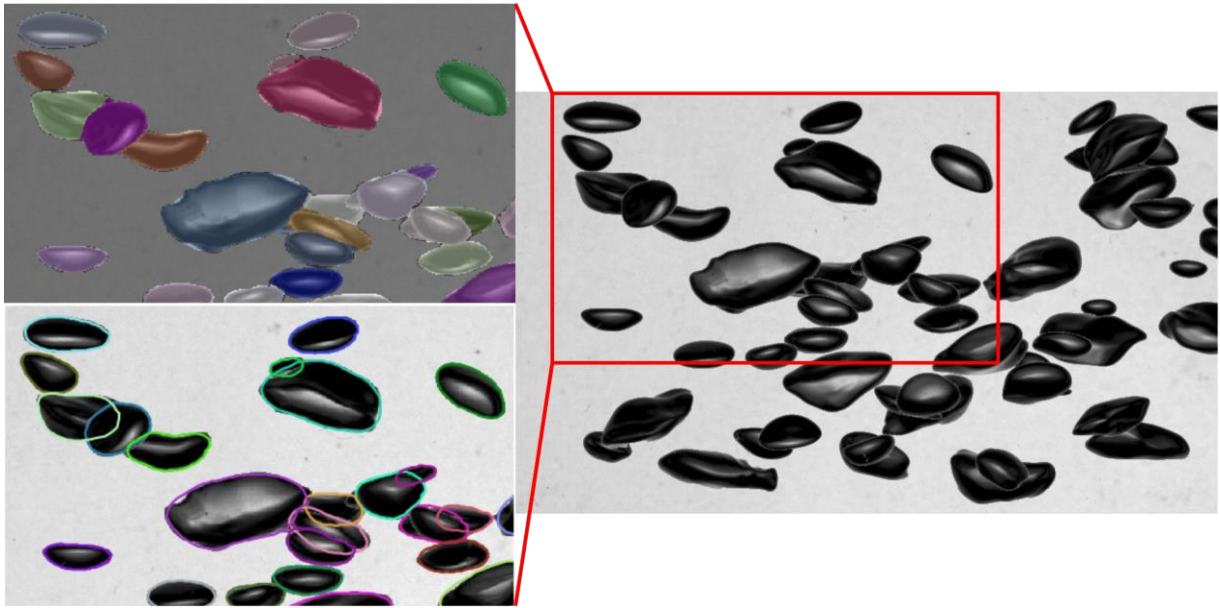

Figure 8: Generated example image for gas volume fraction of 5 % (right) with the segmentation mask from StarDist+UNet (top left) and the overlapping outlines corrected with the RDC method (bottom left).

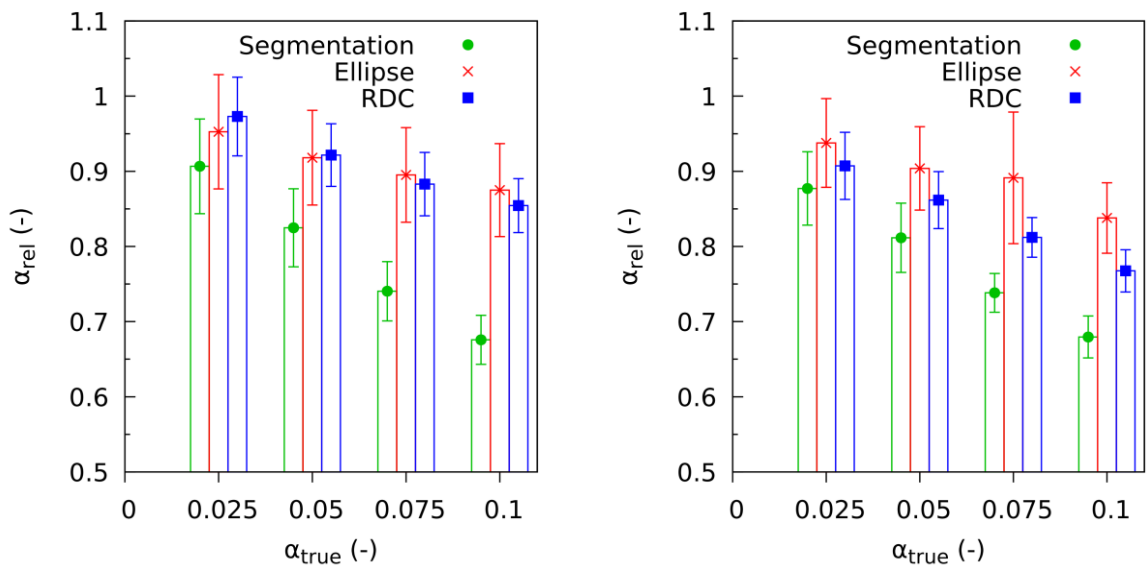

Figure 9: Gas volume fraction results with respect to the ground truth gas volume fraction using ideal segmentation (left) and the combined SD+UNet segmentation (right).

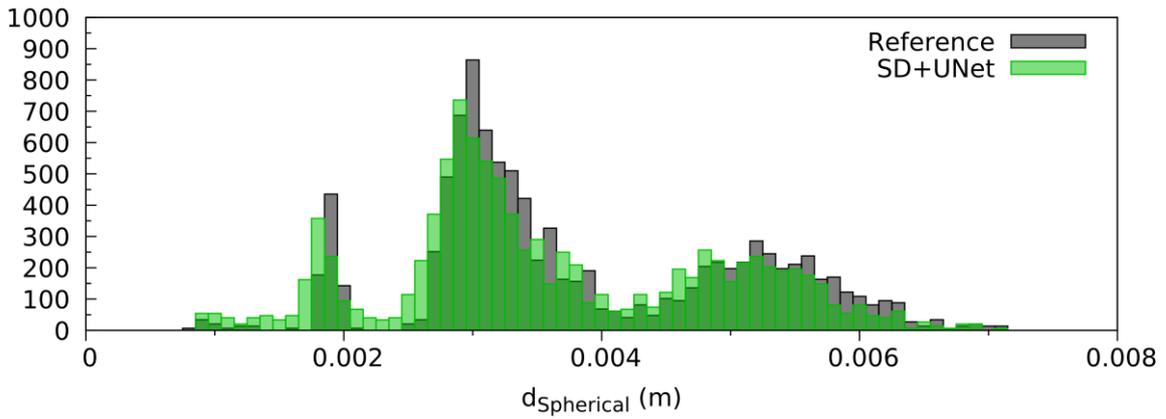

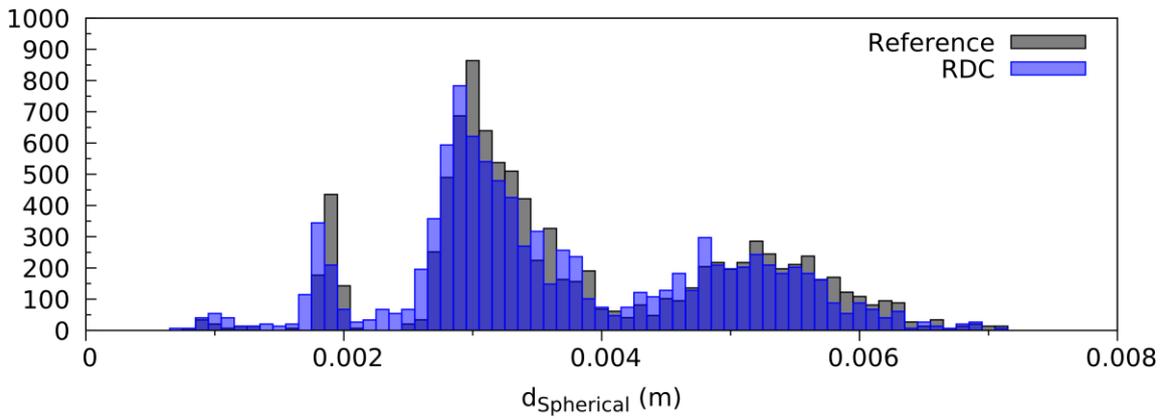

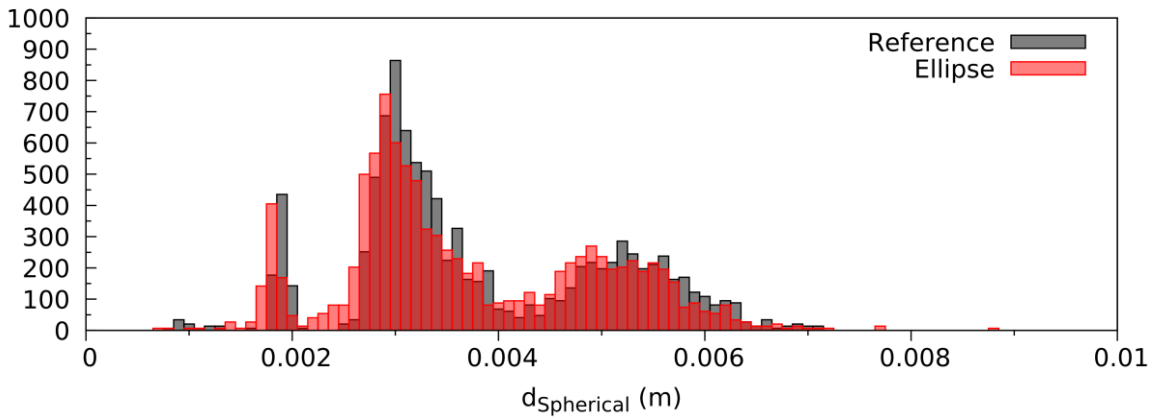

Figure 10: Bubble size histograms for the test case at 2.5 % gas volume fraction. Note that a different abscissa is used for the ellipse fitting method due to some much larger prediction.

Figure 9 shows the error of the predicted gas volume fractions, where $\alpha_{rel\_error} = \frac{\alpha_{predicted}}{\alpha_{reference}}$, using the known ideal segmentation (left), i.e. the visible bubble parts, and using the segmentation predicted by SD+UNet (right) as well as the results when combined with the two hidden part reconstruction methods. The error bars denote the standard deviation for the 50 images of each case. As expected, using only the segmentation mask underpredicts the gas volume fraction due to the missing volume of the hidden parts. Already for the lowest tested gas volume fraction this results in an underprediction of about 9 %. When using the SD+UNet prediction, this error is only slightly higher with about 12 %. With increasing gas volume fraction, this difference even decreases. However, some bubbles are still incorrectly predicted with the SD+UNet method, e.g. some bubbles are missed or multiple bubbles are predicted where only one exists. This becomes important for the hidden part reconstruction. While the RDC method underestimates the gas volume fraction of about 3 % using the ideal segmentation, the error increases to about 9 % when using the SD+UNet prediction. With increasing gas volume fraction, this discrepancy increases showing that the RDC method strongly relies on a correct segmentation. The ellipse fitting method is not that dependent on the segmentation and shows the smallest deviation to the ground truth when combined with the SD+UNet prediction, while, however, showing the highest standard deviation. A deeper look into the underlying bubble size histograms for all images of this first case (shown in Figure 10) reveals that the RDC method better predicts the correct bubble size than without correction as well as with the ellipse fitting method. Note that for the ellipse fitting method a different abscissa is used due to the mentioned fact, that some much larger bubbles are predicted. This is also the reason for the smaller deviation to the ground truth gas volume fraction, as these false larger bubbles correspondingly increase the gas volume fraction. However, the RDC method works better on regular (ellipsoidal) shaped bubbles than on irregular (wobbling) shaped bubbles. This trend continuous for the test cases with higher gas volume fractions as can be seen in the corresponding bubble size histograms shown in Appendix C.

## 4. Conclusion

In this work, we tested the use of different AI-based methods for the task of segmenting and reconstructing overlapping bubbles in bubbly flow images. In particular, we have implemented and tested three different CNN's, namely a combination of two slightly adapted UNet's as well as the two open-source methods StarDist and Mask-RCNN. In general, all three methods are capable to detect bubbles under eased conditions, namely a proper illumination that clearly reveals intersections of overlapping bubbles, rather regular bubble shape and a manageable number of overlapping bubble segments. When these conditions are not met, the Double UNet approach fails to correctly detect the bubbles, while the StarDist and Mask R-CNN methods are more robust under conditions that are more difficult. The StarDist method performs best in identifying bubbles under various image conditions, but slightly misses the correct outline of the bubbles. We were able to improve the latter drawback by combining the StarDist results with the general foreground-background mask provided by one of the two UNets in a postprocessing step. The Mask R-CNN shows an equally good performance as the StarDist method, which underlines the capability of both methods to detect overlapping bubbles in images. However, StarDist shows better computational performance with respect tin terms of a faster training and inference time in comparison to Mask R-CNN.

In order to further increase the accuracy with respect to determining correct bubble sizes and gas volume fractions, we tested two methods to reconstruct the hidden part of partly occluded bubbles. The first method tries to fit an ellipse with the boundary pixels of a segmentation instance, while the second method, called Radial Distance Correction (RDC), is based on a Neural Network that corrects radial distances from the center of the instance to the occluded part. Here, the second method provides more robust results, as the ellipse fitting method occasionally generates far too large predictions. This is further demonstrated in a final combined validation, in which we apply the StarDist+UNet method together with the two methods to reconstruct the hidden bubble part on synthetic images with known ground truth bubbles.

Even though we demonstrate that satisfactory results with respect to bubble size distribution and gas volume fraction can be achieved, the relative error for the latter is still not in a reasonable range when investigating cases with gas fraction above 5 %. This can be improved by generating more training data of such cases. Since labelling images with a high number of overlapping bubbles is cumbersome, the use of generative models like GAN's might be beneficial for this. Furthermore, a CNN model that already proposes overlapping instances could be advantageous for the task of identifying overlapping bubbles, i.e. the multistar approach, which is an extension of the StarDist approach that focuses on the topic of overlapping instances (Walter et al., 2020). Also the use of image sequences could potentially improve the segmentation performance, as hidden bubbles may be better visible at an earlier or later moment of a sequence.

Although only bubble column experiments without an imposed liquid flow are considered in the present work, the training data still should reflect cases with background flow to some extent. Specifically the bubble appearance in the images defines the detectability with a trained algorithm, where the appearance depends on the image condition (e.g. illumination, pixel size, blurriness) as well as on the bubble shape. In this context, a liquid flow field can modify the latter, but only for higher turbulence intensities (Masuk et al., 2021), so that also low to moderate liquid background flows should be reflected in the training data. This, however, needs further investigation and otherwise an adaption of the training data set.

In order to use our data and/or models, we provide two different open-access repositories. The labelled training data and the synthetic test images for the combined validation can be found under (Hessenkemper et al., 2022a). The code to train the UNet's and to apply the combined methods on bubbly flow images as well as the trained UNet, StarDist and RDC models can be accessed under (Hessenkemper et al., 2022b). To train custom StarDist models we propose to access the well-documented original StarDist repository (https://github.com/stardist/stardist). The same applies for the Mask R-CNN algorithm, for which we propose to use the BubMask repository (https://github.com/ywflow/BubMask) as introduced by Kim and Park, 2021.

# Appendix A

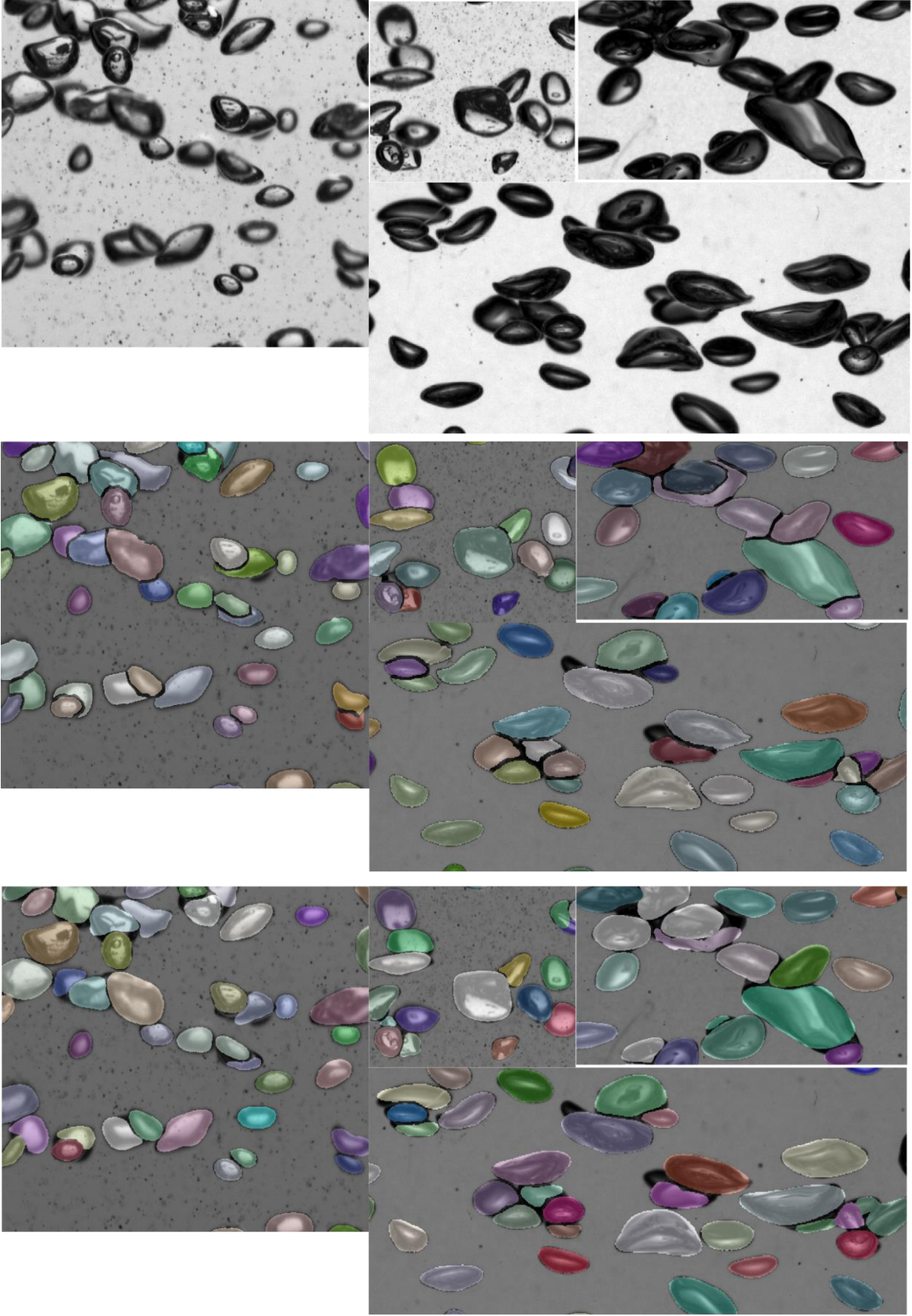

Figure 11: Examples from the validation data set. First row: Original images; Second row: Double UNet prediction; Third row: StarDist prediction. Different colors indicate different detected bubble instances.

Appendix B

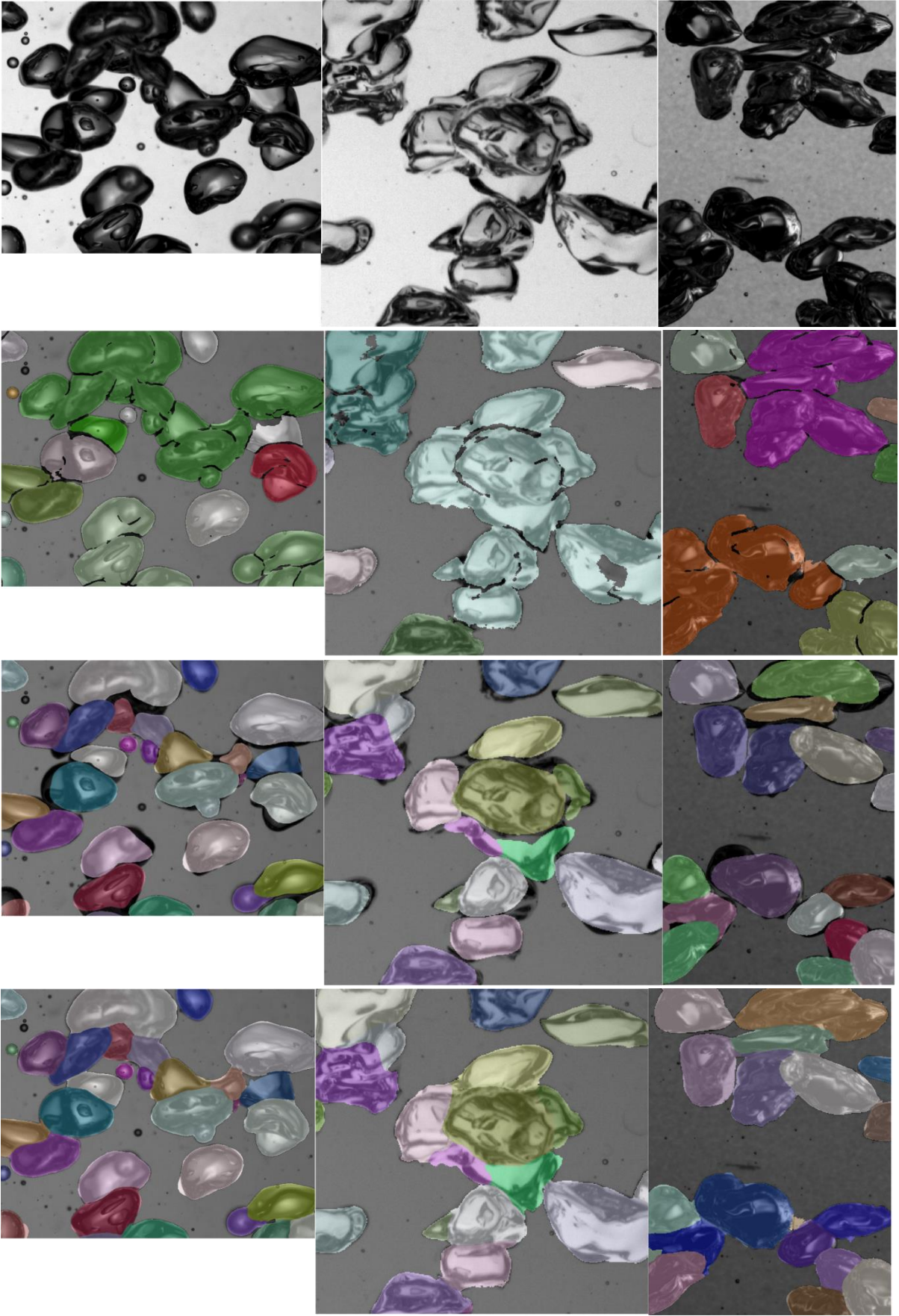

Figure 12: Examples from the test data set. First row: Original images; Second row: Double UNet prediction; Third row: StarDist prediction; Fourth row: Combined prediction.

# Appendix C

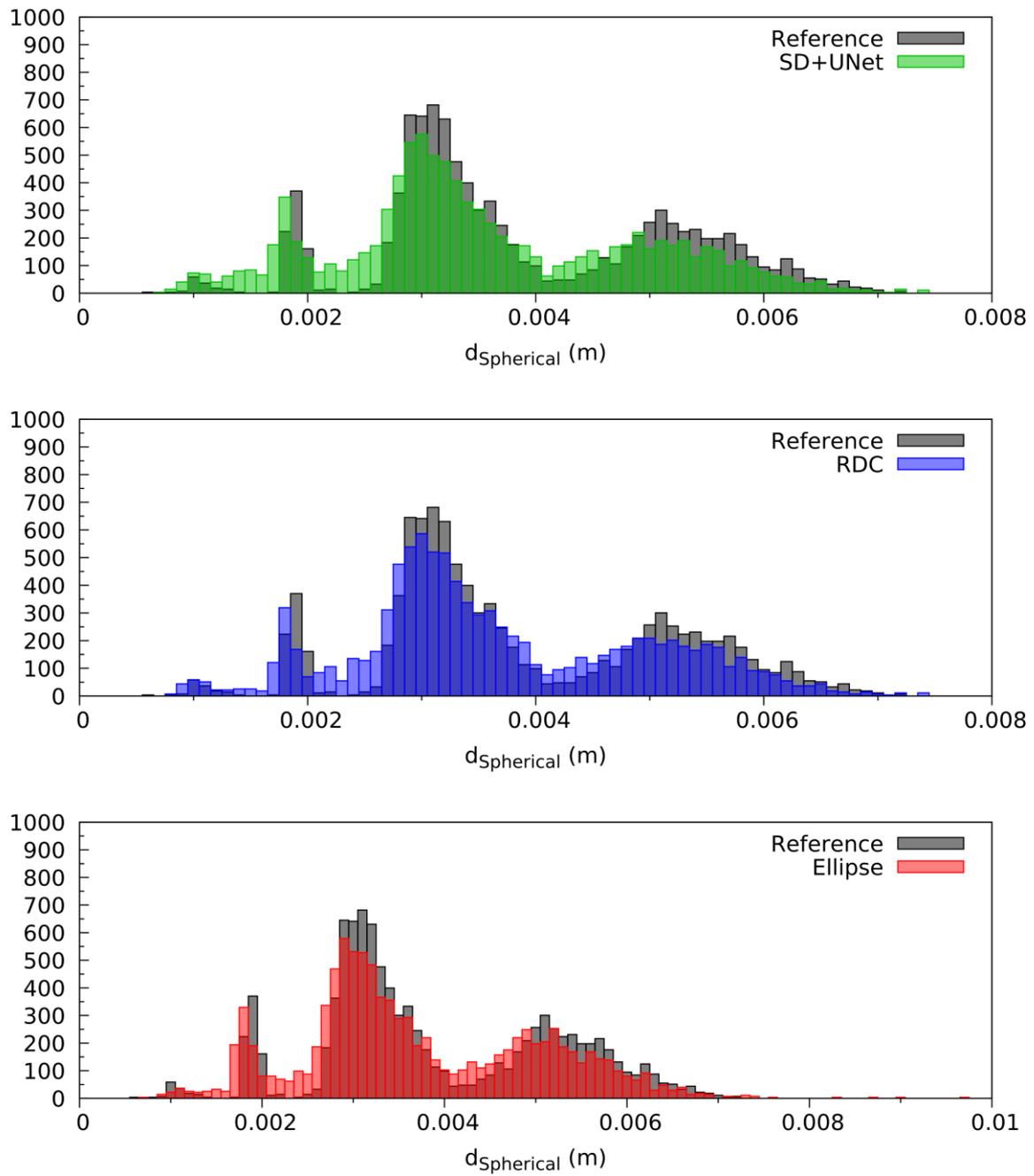

Figure 13: Bubble size histograms for the test case at 5 % gas fraction.

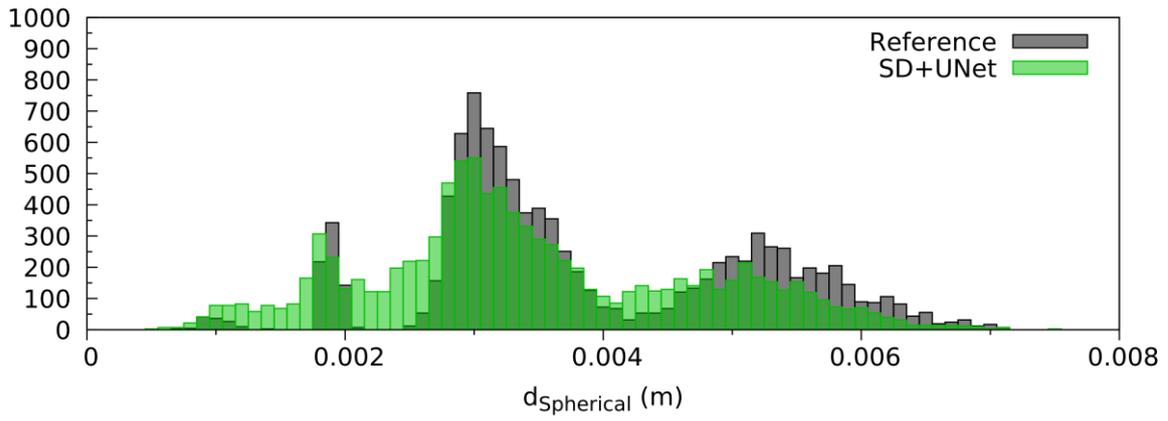
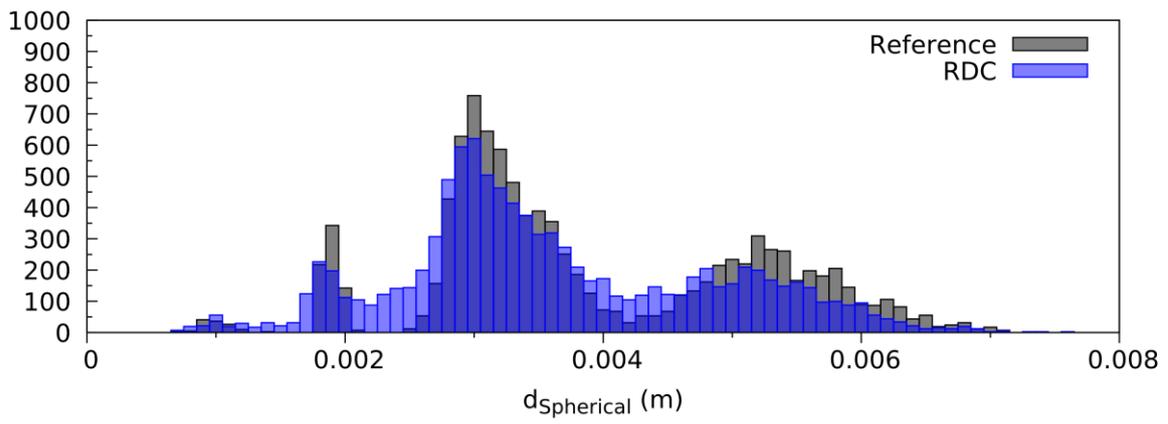
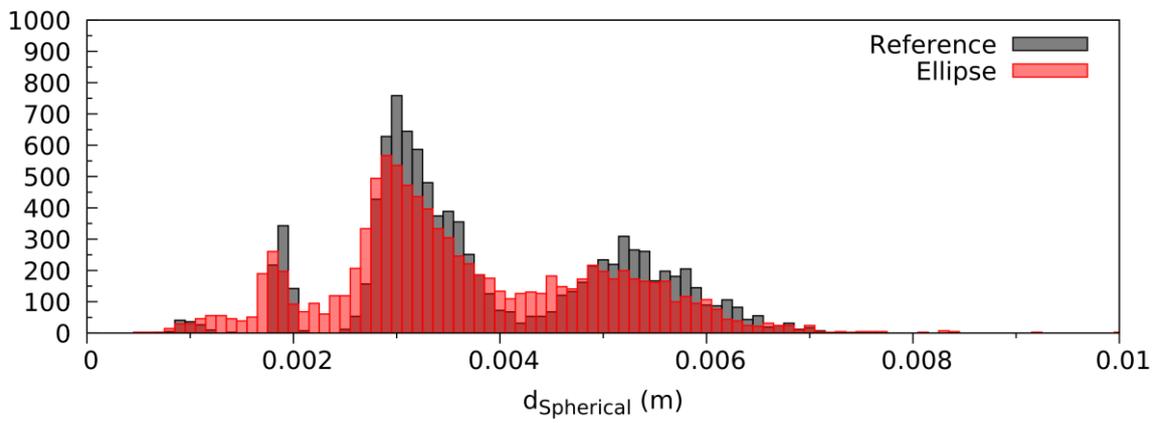

Figure 14: Bubble size histograms for the test case at 7.5 % gas fraction.

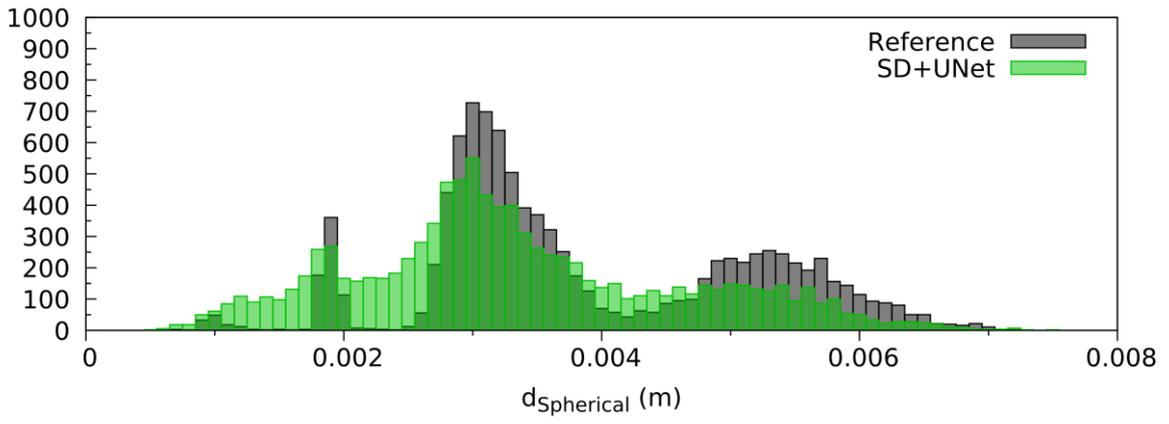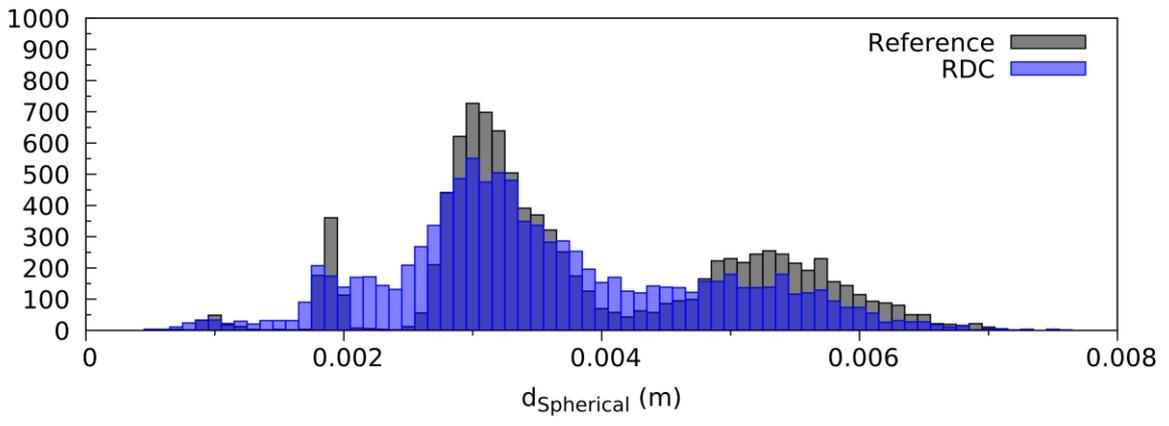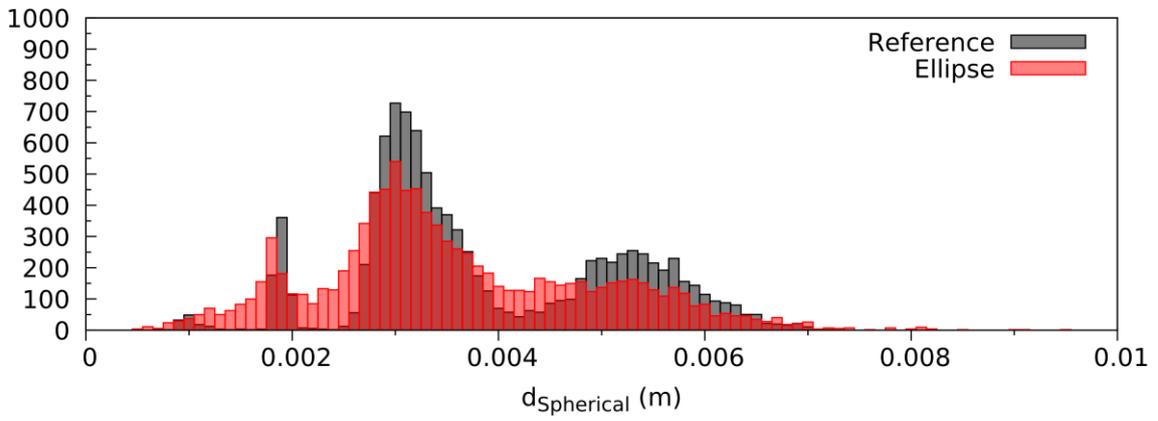

Figure 15: Bubble size histograms for the test case at 10 % gas fraction.